%% file: acl_latex.tex
\pdfoutput=1

\documentclass[11pt]{article}

\usepackage[final]{acl}

\usepackage{times}
\usepackage{latexsym}

\usepackage[T1]{fontenc}

\usepackage[utf8]{inputenc}

\usepackage{microtype}

\usepackage{inconsolata}
\usepackage{graphicx}
\usepackage{stfloats}
\usepackage{CJKutf8}
\usepackage{enumitem}
\usepackage{amsmath}
\usepackage{makecell}
\usepackage{pgfplots}
\pgfplotsset{compat=1.18}
\usepackage{subfigure}
\usepackage[ruled, lined, linesnumbered, commentsnumbered, noend]{algorithm2e}
\usepackage{multicol}
\usepackage{multirow}
\usepackage{booktabs}
\title{PsFuture: A  Pseudo-Future-based Zero-Shot Adaptive Policy for Simultaneous Machine Translation}

\author{Libo Zhao$^{1,2}$,
Jing Li$^{2,3}$\thanks{\,\, Corresponding author.},
Ziqian Zeng$^{1}$\footnotemark[1]\\ 
$^{1}$Shien-Ming Wu School of Intelligent Engineering, South China University of Technology 
\\
$^{2}$Department of Computing, Hong Kong Polytechnic University \\
$^{3}$Research Centre for Data Science \& Artificial Intelligence, Hong Kong Polytechnic University \\
\texttt{wilbzhao@mail.scut.edu.cn,
jing-amelia.li@polyu.edu.hk,
zqzeng@scut.edu.cn}
}

\begin{document}
\begin{CJK}{UTF8}{gbsn}
\maketitle
\begin{abstract}
Simultaneous Machine Translation (SiMT) requires target tokens to be generated in real-time as streaming source tokens are consumed. 
Traditional approaches to SiMT typically require sophisticated architectures and extensive parameter configurations for training adaptive read/write policies, which in turn demand considerable computational power and memory.
We propose PsFuture, the first zero-shot adaptive read/write policy for SiMT, enabling the translation model to independently determine read/write actions without the necessity for additional training. Furthermore, we introduce a novel training strategy, Prefix-to-Full (P2F), specifically tailored to adjust offline translation models for SiMT applications, exploiting the advantages of the bidirectional attention mechanism inherent in offline models. 
Experiments across multiple benchmarks demonstrate that our zero-shot policy attains performance on par with strong baselines and the P2F method can further enhance performance, achieving an outstanding trade-off between translation quality and latency.\footnote{The code is available at \url{https://github.com/lbzhao970/PsFuture}}
\end{abstract}

\begin{figure*}[t]
    \centering
    \includegraphics[width=0.9\textwidth]{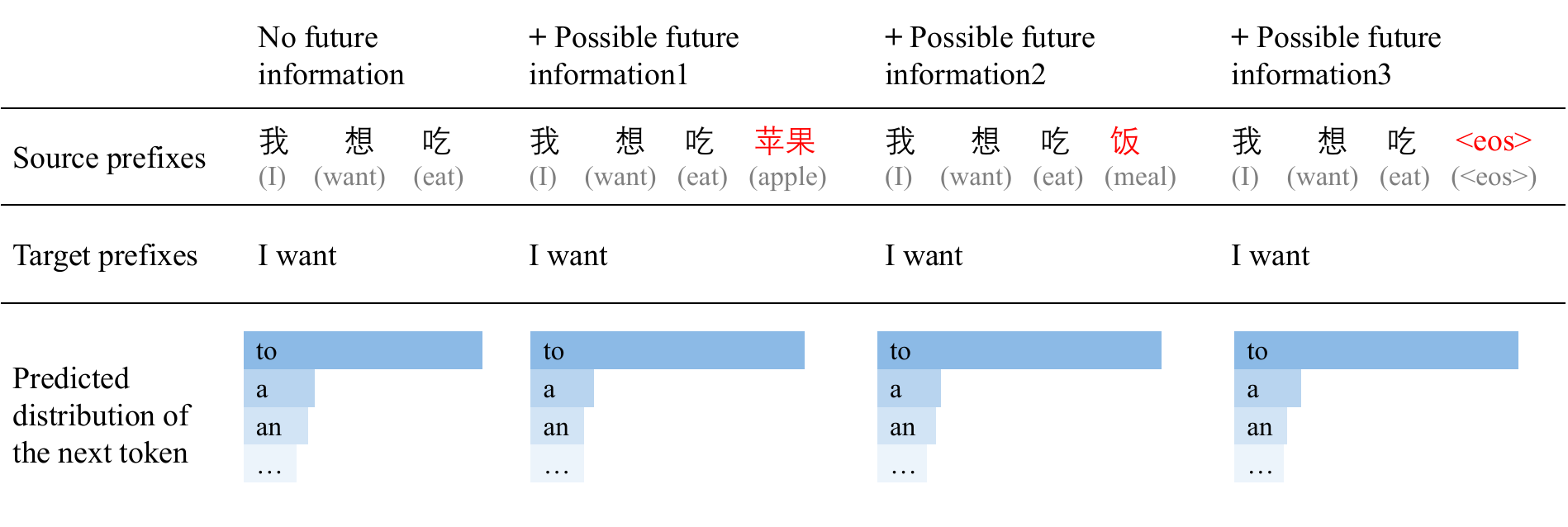}
    \caption{An Zh$\to$En example demonstrating an ideal timing for predicting the next token "to". Even when provided with additional possible future information, the probability distribution of the predicted next token does not change significantly, remaining dominated by the token "to". Therefore, based on the current source prefix "我想吃" and the current target prefix "I want," a write operation can be executed to predict the next token as "to".
    }
    \label{fig:intro}
\end{figure*}
\section{Introduction}
Simultaneous Machine Translation (SiMT) \cite{gu-etal-2017-learning} is required to generate target tokens concurrently as it processes incoming source tokens. Differing from traditional machine translation (MT) methods \cite{bahdanau2015neural,vaswani2017attention,pang2024rethinking} that access the full source text, SiMT necessitates a read/write (R/W) policy to decide between emitting target tokens or awaiting more source input, coupled with the ability to translate from source prefixes to target prefixes (P2P) \cite{ma2018stacl}. Typically, the read/write policy is integrated with the translation mechanism: either employing a fixed wait-$k$ policy alongside a corresponding translation model \cite{ma2018stacl,elbayad20waitk,zhang_future-guided_2021}, or utilizing an adaptive policy \cite{gu-etal-2017-learning,dalvi-etal-2018-incremental,zheng-etal-2019-simpler,zheng2020simultaneous,Ma2020Monotonic,zhang2022reducing,guo2023learning,zhao2024dap,chen2024divergence} that dynamically adjusts read/write decisions based on the context, in conjunction with a model trained to translate policy-defined prefixes. This adaptive method has led to superior performance \cite{ITST,zhang2023hidden}, yet it demands specialized architectural solutions and multitask learning frameworks for concurrent training of the closely linked adaptive policy and translation model, complicating component optimization and increasing computational demands.

On the other hand, DaP-SiMT \cite{zhao-etal-2023-adaptive} introduces a novel framework that separates the adaptive read/write policies from the translation model, offering greater versatility in simultaneous translation. This approach demonstrates that translation models, when directed by an effective adaptive read/write policy, even if initially trained on fixed policies, can balance quality and latency well, achieving state-of-the-art (SOTA) outcomes. However, akin to other adaptive policies, it requires intricate designs and a significant parameter set for training the adaptive read/write policy, often demanding substantial computational resources and memory.

We introduce PsFuture, a zero-shot adaptive read/write policy based on pseudo-future information. This policy utilizes the inherent capabilities of the translation model itself to make read/write decisions without additional training.
Similar to the policy in DaP-SiMT~\cite{zhao-etal-2023-adaptive}, we draw inspiration from human simultaneous translation \cite{Al-Khanji2000compensatory,liu2008experts}, where interpreters shift from listening to translating upon anticipating that further future words would not impact their current decisions. 
As illustrated in Figure~\ref{fig:intro}, this behavior implies a minor divergence between translation predictions based on partial versus more complete source context.
However, in simultaneous translation tasks, previewing future source information is not feasible. Our method, PsFuture, overcomes this by utilizing pseudo-future information, which is a token suffix in the source language that can be fixed or dynamically predicted by language models. By quantifying the divergence between the predicted next target token distributions with or without pseudo-future information, and comparing it to a predefined threshold, a read/write decision can be made. 

The proposed PsFuture method can be directly applied to most existing simultaneous translation models, such as the multi-path wait-$k$ model, which demonstrates superior performance when directed by effective adaptive policies~\cite{zhao-etal-2023-adaptive}.
Additionally, we investigate the application of the PsFuture method to offline translation models. Previous SiMT models~\cite{elbayad20waitk,ITST} conventionally employ a unidirectional attention encoder with tailored masked-cross-attention for prefix-to-prefix training. This approach, while efficient, limits the model's ability to extract features, making it less adept in high-latency scenarios compared to offline models that utilize bidirectional attention mechanisms. To leverage the benefits of bidirectional attention in SiMT, we introduce a novel and effective training technique, Prefix-to-Full (P2F), designed to enhance the performance of offline translation models under diverse latency conditions.
Our main contributions can be summarized as follows.
\begin{enumerate}[topsep=-0.5pt,itemsep=-0.5ex,partopsep=1ex,parsep=1ex]
\item 
We propose the first zero-shot adaptive read/write policy in SiMT, PsFuture, which utilizes the inherent capabilities of the translation model to make read/write decisions without any additional training. To our knowledge, PsFuture is the only adaptive method in the current SiMT field that offers such flexibility.

\item 
We present an effective training technique, Prefix-to-Full (P2F) to enhance the performance of offline translation models under diverse latency conditions.

\item 
Experiments across multiple benchmarks demonstrate that our zero-shot policy attains performance on par with strong baselines and achieves an outstanding accuracy-latency balance.

\end{enumerate}

\section{Related Work}

SiMT policies are broadly categorized into fixed and adaptive schemes. Fixed policies \cite{ma2018stacl,elbayad20waitk,zhang_future-guided_2021} execute read/write actions following predefined rules, such as the wait-$k$ policy \cite{ma2018stacl}, which after reading $k$ source tokens, alternates between reading and writing one token. Conversely, adaptive policies dynamically determine read/write actions based on the evolving source and target context, enhancing the balance between translation accuracy and latency. 

Adaptive approaches employ methods like reinforcement learning within a Neural Machine Translation (NMT) framework \cite{gu-etal-2017-learning}, incremental decoding for variable target token output \cite{dalvi-etal-2018-incremental}, and attention-based methods \cite{arivazhagan2019monotonic,Ma2020Monotonic}. 
Additionally, the wait-info policy \cite{zhang_wait-info_2022} and ITST \cite{ITST}  quantify the waiting latency
and information weight respectively for adaptive policy formulation.
HMT~\cite{zhang2023hidden} optimizes read/write decisions by enhancing the target sequence's marginal likelihood across various translation initiation points. \citet{kim2023enhanced} employs a word-level policy to enhance SiMT. 
Furthermore, \citet{ma2023non} introduces a non-autoregressive streaming Transformer (NAST) to mitigate the challenges of nonmonotonicity and source-information leakage present in conventional autoregressive SiMT frameworks.  \citet{guo2023simultaneous}
propose to provide a tailored reference for the improvement of SiMT model training.

Sharing a similar inspiration with PsFuture, DaP-SiMT \cite{zhao-etal-2023-adaptive} autonomously generates read/write supervisions by leveraging future information divergence for training a decision-making network. In contrast, our approach harnesses the model's inherent translation capability to attain an immediate, zero-shot read/write policy.
\begin{figure*}[t]
    \centering
    \includegraphics[width=0.9\textwidth]{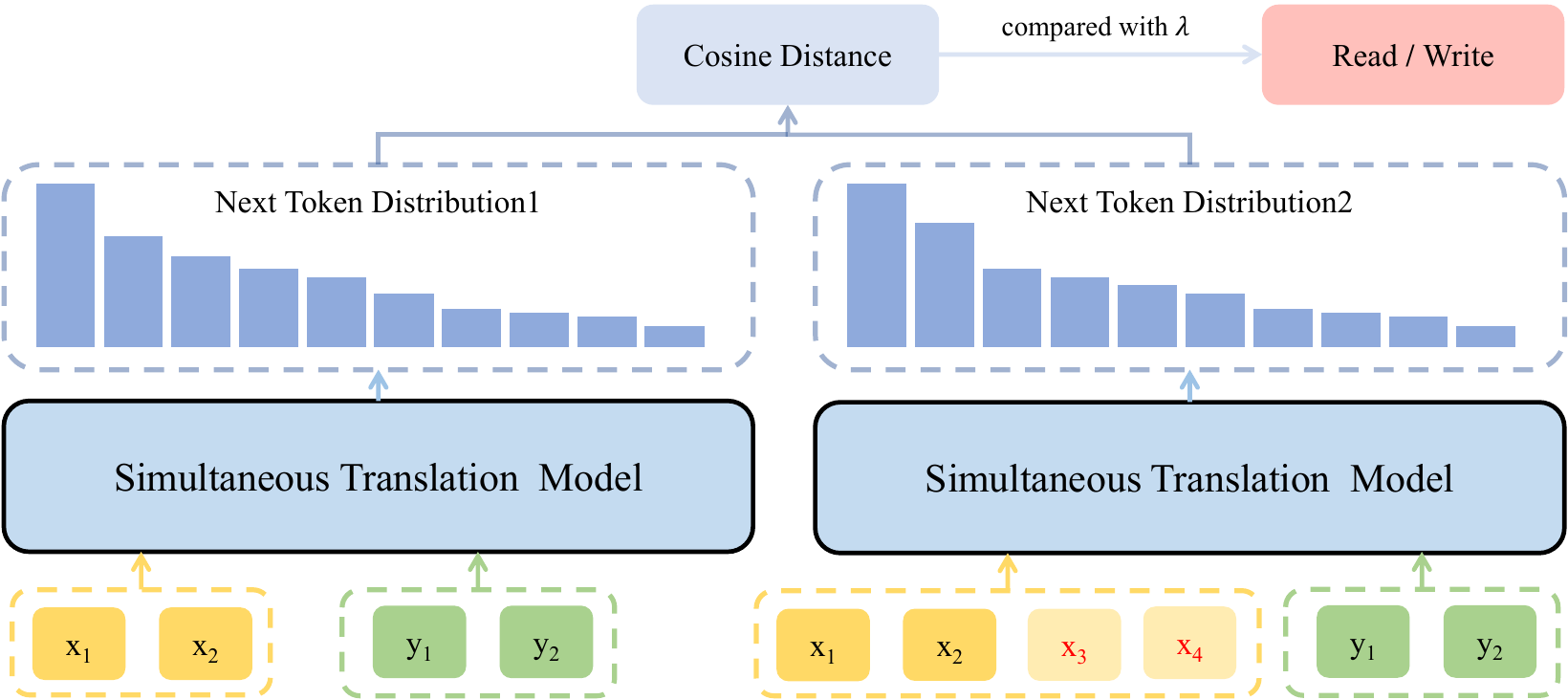}
    \caption{An overall schematic of the PsFuture policy. Based on the current source prefixes ${(x_1, x_2)}$, target prefixes ${(y_1, y_2)}$, and pseudo future information ${(x_3, x_4)}$ (tokens highlighted in red), the simultaneous translation model can directly perform adaptive read/write decisions.}
    \label{fig:psfuture_overall}
\end{figure*}
\section{Preliminary}
\label{sec:prelim}

\subsection{Full-sentence MT and SiMT}

In full sentence translation tasks, an encoder-decoder architecture like the Transformer \cite{vaswani2017attention} transforms a translation pair $\mathbf{x}=(x_1,x_2,...,x_N)$ and $\mathbf{y}=(y_1,y_2,...,y_T)$ by encoding $\mathbf{x}$ into latent representations, followed by the autoregressive generation of target tokens from these representations. 
Generally, the model is optimized by minimizing the cross-entropy loss.
\begin{equation}\label{eq:mt_loss}
\mathcal{L}_{\textnormal{mt}}=-\sum\nolimits_{t=1}^{T} \log p\left(y_{t} \mid \mathbf{x}, \mathbf{y}_{<t} \right)
\end{equation}

For Simultaneous Machine Translation (SiMT), where $g(t)$ denotes a monotonic non-decreasing function indicating the end timestamp of the source prefix required to produce the $t$-th target token, the objective function for SiMT can be adapted as follows,
\begin{equation}\label{eq:simt_loss}
\mathcal{L}_{\textnormal{simt}}=-\sum\nolimits_{t=1}^{T} \log p\left(y_{t} \mid \mathbf{x}_{\leq g(t)}, \mathbf{y}_{<t} \right) .
\end{equation}

\subsection{Wait-k Policy and Multi-Path Wait-k}

\textbf{Wait-$k$ policy} \cite{ma2018stacl}, the most widely used fixed policy, starts by reading $k$ source tokens and then alternates between WRITE and READ action. 
The function $g(t)$ for the wait-$k$ policy can be formally calculated as,
\begin{equation}\label{eq:waitk_gt}
g(t;k) = \min\{t+k-1,N\} .
\end{equation}

\noindent \textbf{Multi-path Wait-$k$} \cite{elbayad20waitk} is an efficient technique for wait-$k$ training. 
It randomly samples different $k$ values between batches during model optimization. 
By employing a unidirectional attention encoder with a tailored upper triangular masked cross-attention mechanism, the multi-path wait-$k$ model achieves efficient prefix-to-prefix training. \citet{zhao-etal-2023-adaptive} demonstrates that the multi-path wait-$k$ model can attain SOTA performance under the guidance of effective adaptive policies.

\section{Method}
\label{sec:method}

\subsection{The Pseudo-Future-based Zero-Shot Adaptive Policy}\label{sec:psfuture method}
In simultaneous translation, skilled human translators execute read/write decisions grounded in the evolving contexts of source and target texts. Conceptualizing a well-trained translation model as an intelligent agent like a human, our objective is to delineate a zero-shot adaptive read/write policy. This approach enables decision-making based purely on the model's inherent linguistic comprehension and translation proficiency, facilitating adaptive policies without necessitating further training.

Zooming in on the details of the read/write decision-making process, 
interpreters transition from listening to translating when they have acquired sufficient source context $\mathbf{x}_{\leq g(t)}$ to decide on extending the partial translation $\mathbf{y}_{<t}$ with the next target word $y_t$. This decision is based on the anticipation that additional source information will not alter their current translation choice, which implies a slight divergence $\mathbf{D}\left(\mathbf{p}^{\textnormal{part}}_t, \mathbf{p}^{\textnormal{more}}_t \right)$ between the interpreters' estimation of the translation distribution with partial source context $\mathbf{p}^{\textnormal{part}}_t$, and the translation distribution considering the more complete source context $\mathbf{p}^{\textnormal{more}}_t$. Interpreters opt to wait for more source words if this divergence becomes substantial.
\begin{align}
    \mathbf{p}^{\textnormal{part}}_t &= p(y_t=\cdot | \mathbf{x}_{\leq g(t)}, \mathbf{y}_{<t}) \label{eq:y_dist_on_prefix} \\
    \mathbf{p}^{\textnormal{more}}_t &= p(y_t=\cdot| \mathbf{x}_\textnormal{more}, \mathbf{y}_{<t}) , \label{eq:y_dist_on_full}
\end{align}  
\noindent where $\mathbf{x}_\textnormal{more}$ represents
 the more complete source context by appending additional source tokens $(x_{g(t)+1},x_{g(t)+2},...)$ to the current source texts $\mathbf{x}_{\leq g(t)}$ and the distributions can be computed by any SiMT translation models.

However, previewing future source information is not feasible during inferring in simultaneous translation. Our proposed PsFuture method, as the name implies, overcomes this by utilizing pseudo-future information $\mathbf{x}_\textnormal{ps-suffix}$, which is a token suffix in the source language.
It should be noted that pseudo-future information here does not merely refer to the predicted next few source tokens adhering to human natural language patterns, but rather a broader concept representing additional information beyond current source input.
When such information minimally impacts the subsequent target token prediction of the translation agent, it indicates low ambiguity in the translating process, which suggests that the translation of the current source prefix remains incomplete, thereby signaling an appropriate moment for a WRITE operation.
Conversely, it indicates an opportune moment for a READ operation.
In this work, we explore various forms of pseudo-future information, including both predefined fixed suffixes and adaptive ones that are dynamically predicted by language models (detailed in Section~\ref{experiment:ps-suffix}). 
 
As shown in Equation~\ref{eq:cosine} and~\ref{eq:y_dist_on_pseudo}, we utilize cosine distance, which has been validated as effective in DaP-SiMT \cite{zhao-etal-2023-adaptive}, to quantify the divergence $\mathbf{D}\left(\mathbf{p}^{\textnormal{part}}_t, \mathbf{p}^{\textnormal{pseudo}}_t \right)$ between the predicted next target token distributions with or without pseudo-future information.
\begin{align}
    \mathbf{D}& \left(\mathbf{p}^{\textnormal{part}}_t, \mathbf{p}^{\textnormal{pseudo}}_t \right) = 1 - \cos\left(\mathbf{p}^{\text{part}}_t, \mathbf{p}^{\text{pseudo}}_t\right) \label{eq:cosine} \\
    & \mathbf{p}^{\textnormal{pseudo}}_t = p(y_t=\cdot| \mathbf{x}_\textnormal{pseudo}, \mathbf{y}_{<t}) , \label{eq:y_dist_on_pseudo} 
\end{align}  
\noindent where $\mathbf{x}_\textnormal{pseudo}$ represents
 the fake complete source context by appending pseudo future information $\mathbf{x}_\textnormal{ps-suffix}$ to the current source texts $\mathbf{x}_{\leq g(t)}$.

 By comparing the divergence value to a predefined threshold $\lambda$, a read/write decision can be made as Equation~\ref{eq:thresholding}. The overall schematic of the PsFuture policy is illustrated in Figure~\ref{fig:psfuture_overall}. 
 \begin{equation}
    \textit{write } \text{if } \mathbf{D}_{t,g(t)} < \lambda,  \text{else } \textit{read} \label{eq:thresholding}
\end{equation}

\begin{figure}[t]
    \centering
    \includegraphics[width=0.45\textwidth]{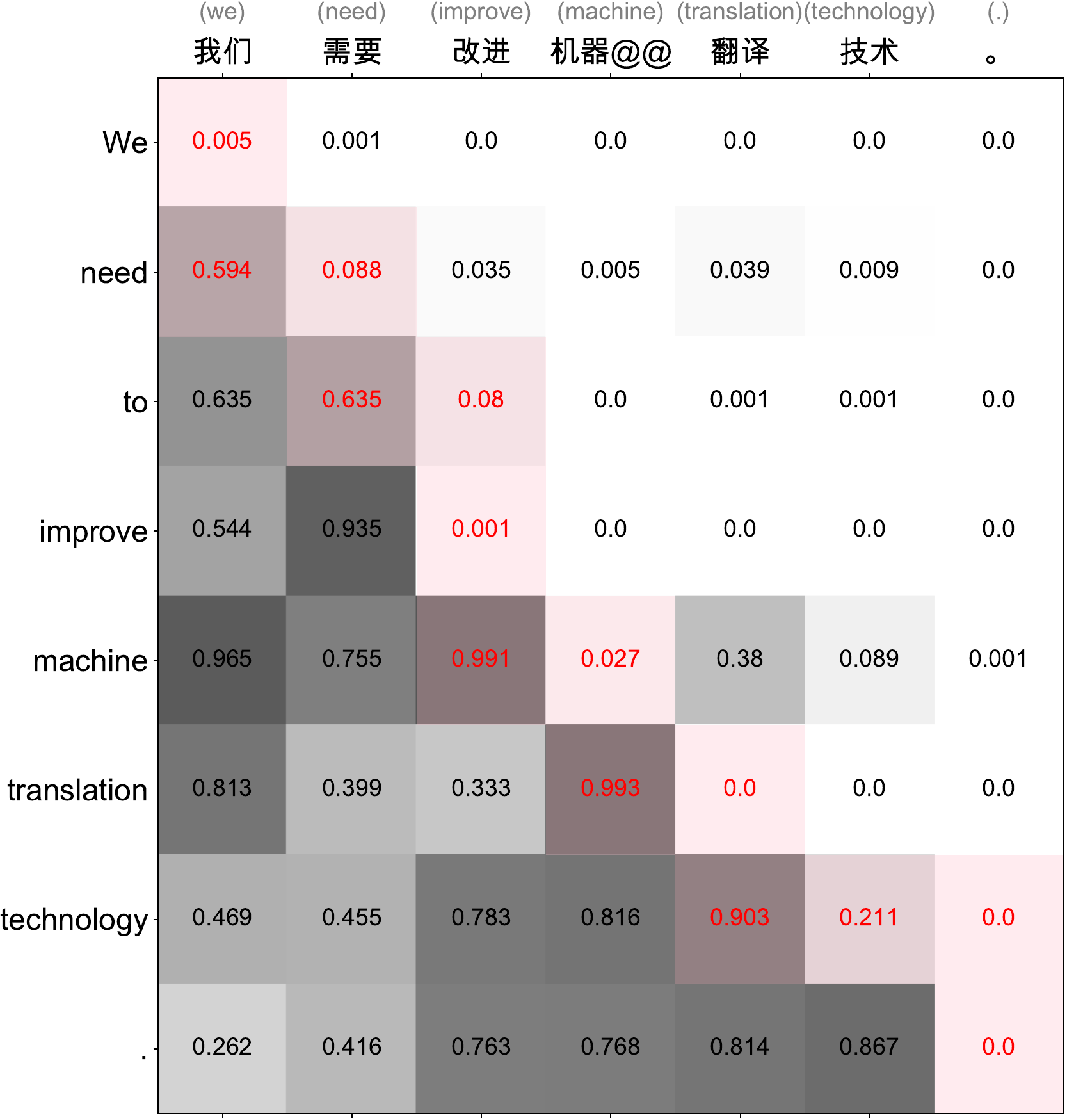}
    \caption{Example of a Zh$\to$En divergence matrix $\mathbf{D}$, where $\mathbf{D}_{t,g(t)}=\mathbf{D}\left( \mathbf{p}^{\textnormal{part}}_t,\mathbf{p}^{\textnormal{pseudo}}_t \right)$. The red elements in the matrix denote a potential read/write path, determined by a predefined threshold $\lambda$ (0.2 in this case).}
    \label{fig:psfuture_matrix}
\end{figure}

Figure~\ref{fig:psfuture_matrix} shows an example divergence matrix based on the PsFuture method and a highlighted read/write path, in which we only employ a ``<eos>'' token as the pseudo-future suffix. It can be observed that comparing with a suitable threshold allows for the easy identification of a potential read/write path.

Following~\cite{zhao-etal-2023-adaptive}, we also introduce another hyperparameter in the read/write decision-making process to limit the maximum number of continuous READ operations for certain languages, thereby enhancing their performance. The inference process is summarized in Algorithm~\ref{alg:rw_policy}.
\begin{table}[b]
\centering
\begin{tabular}{lll}
\hline
& Zh$\to$En& De$\to$En \\
\hline
Standard Transformer &  20.32 & 32.99\\ 
Multi-path Wait-$k$ &  19.45&31.81\\ 
ITST &  19.15 & 31.26\\
\hline
\end{tabular}
\caption{\label{table:offline bleu}
Comparison of case-insensitive BLEU in offline scenario among the standard Transformer model\cite{vaswani2017attention}, multi-path wait-$k$ model\cite{elbayad20waitk} and ITST\cite{ITST}.}
\end{table}
\subsection{The Prefix-to-Full Training Method Adapting Offline Models to SiMT }\label{sec:p2f method}
The PsFuture approach is versatile, compatible with most translation models, including the offline ones\footnote{In this paper, we distinguish standard Transformer models, which employ the bidirectional attention mechanism, as offline translation models, to differentiate them easily from SiMT models that utilize the unidirectional attention mechanism.} (standard Transformer~\cite{vaswani2017attention}). 
Offline translation models have demonstrated substantial potential for simultaneous translation, as evidenced by their efficacy in speech translation~\cite{papi2022does}.
However, the lack of Prefix-to-Prefix (P2P) training in offline models leads to lower translation quality under low-latency conditions compared to SiMT models~\cite{ma2018stacl}. On the other hand, the bidirectional attention mechanism of offline models significantly enhances feature extraction, surpassing the unidirectional attention mechanism typically used in SiMT models to facilitate P2P training. Thus, in high-latency scenarios, offline models usually achieve better translation quality, as shown in Table~\ref{table:offline bleu}.

To harness the benefits of the bidirectional attention mechanism in real-time contexts, we introduces a simple yet effective training strategy for offline translation models named Prefix-to-Full (P2F). This method aims to preserve the model's superior performance in high-latency scenarios while improving its effectiveness in mid-to-low latency situations. The training regimen not only utilizes the conventional translation loss as  Equation~\ref{eq:mt_loss}, but also integrates an innovative loss function, Prefix-to-Full (P2F) loss. P2F loss is designed to translate a source prefix into a complete sentence, with the prefix length $l$ being uniformly distributed and randomly chosen. The overall loss is computed as follows.
\begin{align}\label{eq:final_loss}
 \mathcal{L}_{\textnormal{total}} &=  (1 - \alpha)\mathcal{L}_{\textnormal{mt}} + \alpha\mathcal{L}_{\textnormal{p2f}} \\
\mathcal{L}_{\textnormal{p2f}} &= -\sum\nolimits_{t=1}^{T} \log p\left(y_{t} \mid \mathbf{x}_{\leq l}, \mathbf{y}_{<t} \right) \\
l &\sim \textnormal{Uniform}(L) \\
\alpha &\sim \textnormal{Bernoulli}(r) ,
\end{align}
\noindent where $r$ is a hyperparamete to control the proportion of the P2F loss. $L$ is the candidate set of the prefix length $l$, or more specifically, $L = \{1, 2, ..., |\mathbf{x}|\}$.

The P2F loss endows offline translation models with the capability to translate prefixes. Although translating prefixes into full target sentences increases the risk of hallucinations during the simultaneous translation process, we posit that an effective read/write policy can mitigate such occurrences. For a detailed analysis of experiments on this, please refer to Section~\ref{sec:ablation}.

\begin{table*}
\centering
\begin{tabular}{p{2.5cm}p{3cm}p{3.5cm}p{3.5cm}}
\hline
& Zh$\to$En& De$\to$En & En$\to$Vi \\
\hline
Fixed suffix1 & ``<eos>'' & ``<eos>'' & ``<eos>''\\ 
Fixed suffix2 & ``<unk> <eos>'' & ``<unk> <eos>'' & 
 ``<unk> <eos>''\\ 
Fixed suffix3 & ``… <eos>'' & ``... <eos>'' & ``... <eos>''\\ 
Fixed suffix4 & ``…信息到此中断。 <eos>'' & ``... Die Informationen enden hier. <eos>'' & 
 ``... Information interrupted here. <eos>''\\ 
Random suffix & \multicolumn{3}{c}{\textcolor{blue}{( Sampled Randomly Each Time A Read/Write Decision Occurs)}}\\
Adaptive suffix & \multicolumn{3}{c}{\textcolor{blue}{(Dynamically Generated by Language Models)}}\\
\hline
\end{tabular}
\caption{\label{table:ps-suffix}
The pseudo-future suffixes across various language pairs utilized in this paper.}
\end{table*}
\section{Experiments}


\subsection{Datasets}

\textbf{WMT2022 Zh$\to$En}\footnote{\scriptsize\url{www.statmt.org/wmt22}}. We use a subset with 25M sentence pairs for training\footnote{\scriptsize{The data sources include casia2015, casict2011, casict2015, datum2015, datum2017, neu2017, News Commentary V16, ParaCrawl V9.}}, from which 1500 unique sentence pairs are extracted as the validation set. We first tokenize the Chinese and English data using the Jieba Chinese Segmentation Tool\footnote{\scriptsize\url{https://github.com/fxsjy/jieba}} and Moses\footnote{\scriptsize\url{https://github.com/moses-smt}}, respectively, and then apply BPE with 32000 merge operations. We employ the dev set of 956 sentence pairs from BSTC \cite{zhang2021bstc} as the test set.


\noindent \textbf{WMT15 De$\to$En}\footnote{\scriptsize\url{www.statmt.org/wmt15}}. All 4.5M sentence pairs from this dataset are used for training, and are tokenized using 32K BPE merge operations. We use newstest2013 (3000 sentence pairs) for validation and report results on newstest2015 (2169 sentence pairs).


\noindent \textbf{IWSLT15 En$\to$Vi}\footnote{\scriptsize\url{nlp.stanford.edu/projects/nmt}}. All 133K sentence pairs from this dataset \cite{luong2015stanford}  are used for training. 
We use TED tst2012 (1553 sentence pairs) for validation and TED tst2013 (1268 sentence pairs) as the test set.  Following the settings in \cite{Ma2020Monotonic}, we adopt word-level tokenization and replace rare tokens (frequency $<5$) with <unk>. 
The vocabulary sizes are 17K for English and 7.7K for Vietnamese, respectively.

\input{figs_tex/main_result}
\subsection{Settings}\label{experiment:ps-suffix}
\noindent \textbf{The Pseudo-Future Suffix} .
In this study, we investigate various pseudo-future suffixes, denoted as $\mathbf{x}_\textnormal{ps-suffix}$, as detailed in Table~\ref{table:ps-suffix}.
These suffixes can be divided into two categories: fixed and adaptive. 
A primary criterion for selecting a fixed suffix is its richness in information. For instance, the ``<eos>'' token, often encountered in training translation models, effectively indicates sentence termination. Consequently, all chosen suffixes conclude with ``<eos>'' to guarantee an essential increment of information.




Specifically, the fixed suffixes range from the basic ``<eos>'' token (suffix 1) to more complex structures involving special tokens (``<unk> <eos>'', suffix 2) and natural sentence extensions (suffixes 3 and 4), which simulate ellipsis and ellipsis with signals of information discontinuity. 
We also conduct an experiment with random suffixes to investigate the sensitivity of the PsFuture method to suffix content. These random suffixes consist of four tokens, each randomly selected from the top 200 most frequent tokens in the vocabulary, ensuring adequate information. Furthermore, the suffix is resampled randomly each time a read/write decision occurs.

The adaptive suffix is dynamically generated by large language models, based on the current source prefix for pseudo-future information prediction. For Zh$\to$En, we employ the Chinese-Llama-2-7b model\footnote{\scriptsize\url{https://github.com/LinkSoul-AI/Chinese-Llama-2-7b}}, while the Llama-2-7b-chat~\cite{touvron2023llama} is used for De$\to$En and En$\to$Vi experiments. 
In the main results (\ref{sec:main}), we empirically determine the optimal suffix through performance evaluation. Section~\ref{sec:ablation} delves into the effects of various suffixes on the experimental outcomes, providing a thorough assessment.

\noindent \textbf{The Prefix-to-Full Loss Ratio}. 
The hyperparameter P2F ratio $r$ is employed to control the proportion of the P2F loss.
The most effective configurations are identified as 0.5, 0.8, and 0.5 for Zh$\to$En, De$\to$En, and En$\to$Vi, respectively. Detailed information on the ablation studies concerning hyperparameter $r$ is referred to Section~\ref{sec:ablation}.

\noindent \textbf{Other Settings}.
The proposed PsFuture policy undergoes empirical experiments based on the multi-path wait-$k$ model and the P2F-enhanced offline model as mentioned in Section~\ref{sec:p2f method}, comparing its performance with two leading models in the SiMT domain, ITST \cite{ITST} and DaP-SiMT \cite{zhao-etal-2023-adaptive}.
All our implementations are based on the Transformer \cite{vaswani2017attention} architecture and adapted from the Fairseq Library \cite{ott-etal-2019-fairseq}. 
For the Zh$\to$En experiments, we utilize the transformer big architecture, while the base and small architectures are used for De$\to$En and En$\to$Vi experiments respectively.


For evaluation, following ITST and DaP-SiMT, we report case-insensitive BLEU \cite{bleu} scores to assess translation quality and Average Lagging (AL/token) \cite{ma2018stacl} to measure latency. Regarding the maximum number of continuous read actions in our method, we empirically select the best-performing configurations, which are no constraint, 4, no constraint for Zh$\to$En, De$\to$En, En$\to$Vi respectively. Furthermore, to achieve more robust inference results, the initial length of the source prefix during the real-time translation process is set to 2.

\input{figs_tex/ablation_study1}
\subsection{Main Results}\label{sec:main}

We compare the proposed PsFuture method against previous approaches for three language pairs in Figure~\ref{fig:main_result}. 
PsFuture-W and PsFuture-O refer to the multi-path wait-$k$ model-based PsFuture approach and the offline model (P2F enhanced) based PsFuture method, respectively.

Firstly, the PsFuture-W experiment significantly surpasses traditional multi-path wait-$k$ models, benefiting from the proposed PsFuture policy over the fixed wait-$k$ policy. Notably, the performance of PsFuture-W often matches or exceeds the SiMT leading model ITST, which is specifically trained with a complicated adaptive read/write policy. This highlights the capability of SiMT translation models to make adaptive decisions themselves. Although trained with a fixed strategy, the multi-path wait-$k$ model, when coupled with the zero-shot PsFuture policy, significantly outperforms its counterparts and rivals strong SiMT baselines.

Secondly, the performance of PsFuture-O demonstrates improvements over PsFuture-W to varying extents across all language pairs, especially in the Zh$\to$En experiment where it outdoes the former SiMT SOTA method, DaP-SiMT. 
As anticipated, the offline translation model, endowed with superior feature extraction capabilities, achieves better performance at moderate to high latencies, while the introduction of the Prefix-to-Full loss ensures the model maintains comparable effectiveness at lower latencies.

\section{Analysis}
In this scetion, we aim to provide a detailed examination of the proposed method. Unless otherwise noted, the PsFuture-related experiments are based on the multi-path wait-$k$ model, and the results stem from the Zh$\to$En Transformer-Big model.

\subsection{Effect of the pseudo-future suffix}\label{sec:ablation}
This part investigates the influence of various pseudo-future suffixes (Table~\ref{table:ps-suffix}) on the experiment results. As shown in Figure~\ref{fig:effect of the pseudo future suffix}, the majority of suffixes tested can achieve a desirable equilibrium between translation quality and latency, which showcases the tolerance of the proposed method to the choice of suffixes. 
Through the comparison of various experimental results, it is also feasible to identify specific suffixes for particular language pairs to optimize performance.
Adaptive suffixes, generated by large language models, consistently perform well across various corpora. However, due to a lack of extensive experimentation with different adaptive suffixes, their effectiveness does not surpass that of the best fixed suffixes. We believe that a large-scale exploration of adaptive suffix experiments could potentially yield superior outcomes.

Additionally, it is surprising that the random suffix experiment exhibits unexpectedly strong performance. Although there are fluctuations in specific areas, the overall result is comparable to that of other meticulously crafted suffixes. This indicates that PsFuture's effectiveness is not significantly affected by suffix content. 
This finding indicates that the proposed method possesses a substantial lower bound, emphasizing its robustness and straightforward applicability. These qualities align with the method's key features: simple yet effective.

Furthermore, experiments with ground truth suffixes are conducted to ascertain the upper bound of the PsFuture method. The results indicate that there remains potential for enhancement. Future efforts will focus on incrementally approaching this upper limit by exploring and refining suffixes.


\subsection{Effect of the P2F loss}

Figure~\ref{fig:p2f} illustrates the impact of different Prefix-to-Full (P2F) loss ratios on the performance of our experiments. Setting the P2F ratio $r$ to 0 corresponds to conventional offline translation model training. This configuration, when applied directly to SiMT tasks, yields less than ideal results, especially at lower to medium latencies. Incorporating any level of P2F loss markedly improves performance, effectively tailoring the offline model for SiMT applications. Moreover, the experimental results reveal a noticeable sensitivity to the P2F ratio $r$, indicating that an optimal $r$ can enhance the balance between translation accuracy and latency.
\input{figs_tex/effect_of_p2f}

\subsection{Concerns on Hallucination}\label{sec:hallucination}
\input{figs_tex/hallucilation_rate}
In the PsFuture-O experiment, the additional introduction of the Prefix-to-Full (P2F) loss aims to enhance the model's capability to translate a source prefix into a full target sentence, thereby adapting it for SiMT tasks. However, this approach may increase the risk of hallucinations during the translation process. A hallucination is defined as a generated token that cannot be aligned with any source word. To illustrate this potential issue, we compare the hallucination rate \cite{chen-etal-2021-improving-simultaneous} of hypotheses generated by PsFuture-O with those produced by other methods. The comparative results are depicted in Figure~\ref{fig:hr}.

It is evident that, overall, the PsFuture-O experiment achieves the lowest hallucination rate, surpassing not only the DaP-SiMT and PsFuture-W methods, which rely on the multi-path wait-$k$ model, but also outperforming the meticulously trained ITST model. This indicates that the proposed PsFuture policy effectively mitigates the occurrence of hallucinations during the simultaneous translation inference process.

\section{Conclusion}

In this paper, we propose the first zero-shot adaptive read/write policy for SiMT, PsFuture. 
It empowers the translation model to autonomously decide on read/write actions without requiring additional training and can attain effectiveness on par with previously meticulously trained adaptive policies.
Moreover, we introduce a novel training strategy, Prefix-to-Full (P2F), specifically tailored to adjust offline translation models for SiMT applications, exploiting the benefits of the bidirectional attention mechanism inherent in offline models. 

\section*{Limitations}

 In this work, the proposed PsFuture policy conducts two forward computations for each read/write decision-making, which may increase the total computational load when inferring. However, it's important to note that while other adaptive policy methods may require only one forward computation for each decision, they also necessitate additional computations, which are also not negligible when compared to single forward computing.
Overall, despite the increased computational requirement for inference, the PsFuture method eliminates the need for additional learnable parameters and training to obtain a read/write decision maker, which also significantly reduces computational demands during training.
\section*{Ethics Statement}
After careful review, to the best of our knowledge, we have not violated the \href{https://www.aclweb.org/portal/content/acl-code-ethics}{ACL Ethics Policy}.

\section*{Acknowledgements}
This work was supported by the National Natural Science Foundation of China (No. 62406114), the Guangzhou Basic and Applied Basic Research Foundation (No. 2023A04J1687), the Fundamental Research Funds for the Central Universities (No. 2024ZYGXZR074), the NSFC Young Scientists Fund (No. 62006203), the Research Grants Council of the Hong Kong Special Administrative Region (No. PolyU/25200821), the Innovation and Technology Fund (No. PRP/047/22FX), PolyU Research Centre on Data Science and Artificial Intelligence (No. 1-CE1E) and a gift fund from Microware (No. N-ZDG2).

\bibliography{custom}

\clearpage
\newpage
\appendix
\section{Effect of The Max Continuous READ Constraint} 

Following DaP-SiMT~\cite{zhao-etal-2023-adaptive},  we set a constraint on the maximum consecutive reads allowed during inference, necessitating a write action once this limit is reached. Figure~\ref{fig:max conti read} demonstrates the influence of this hyperparameter on various language pairings. Consistent with DaP-SiMT, we note that this parameter exerts little or even negative impact on the Zh$\to$En and En$\to$Vi experiments, yet it proves substantially advantageous for the De$\to$En pair. Thus, it is advisable to identify the optimal maximum number of continuous reads on the validation set before the practical implementation of this approach.
\input{figs_tex/ablation_study2}

\section{Case Study}

\begin{figure*}[t]
    \centering
    \includegraphics[width=1\textwidth]{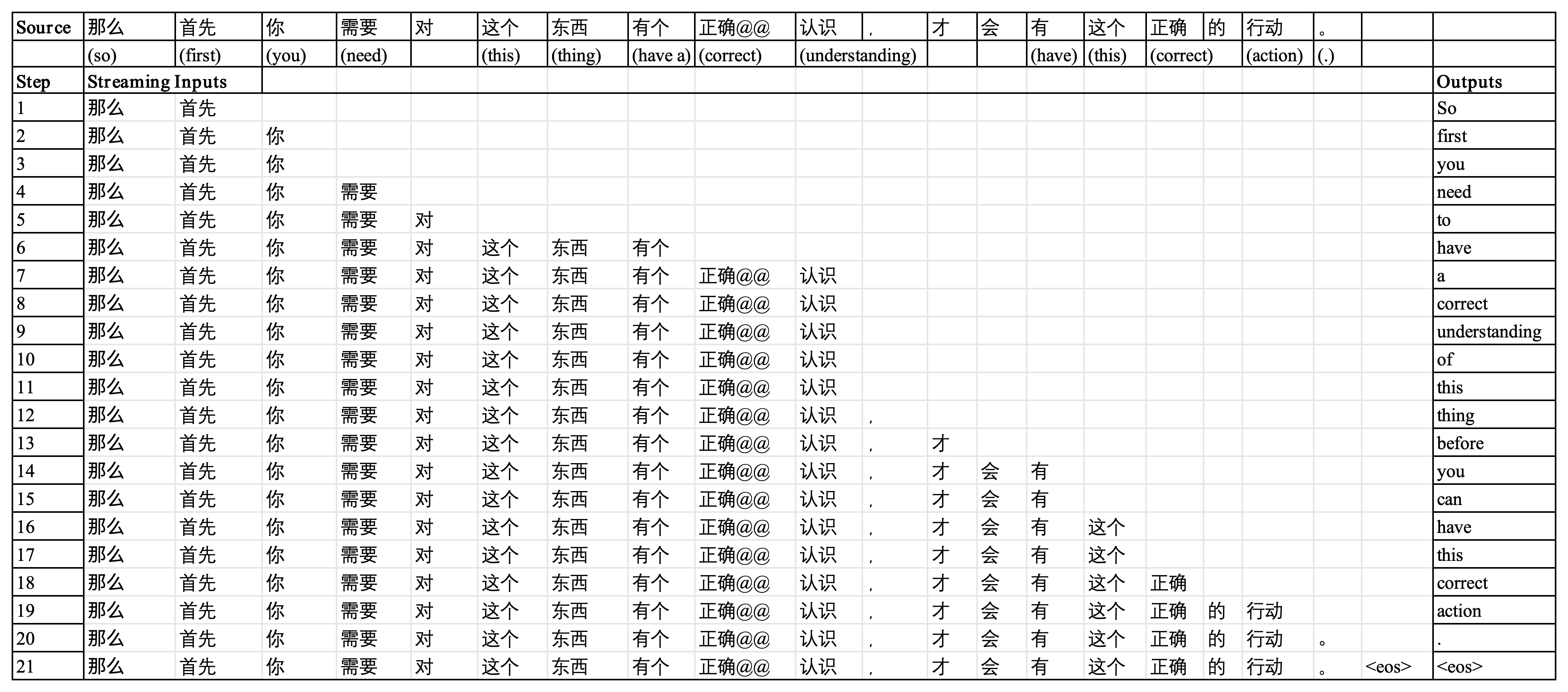}
    \caption{Case No.226 in BSTC Zh$\to$En test set, evaluated with $\lambda$ = 0.08.}
    \label{fig:case_study1}
\end{figure*}

\begin{figure*}[t]
    \centering
    \includegraphics[width=1\textwidth]{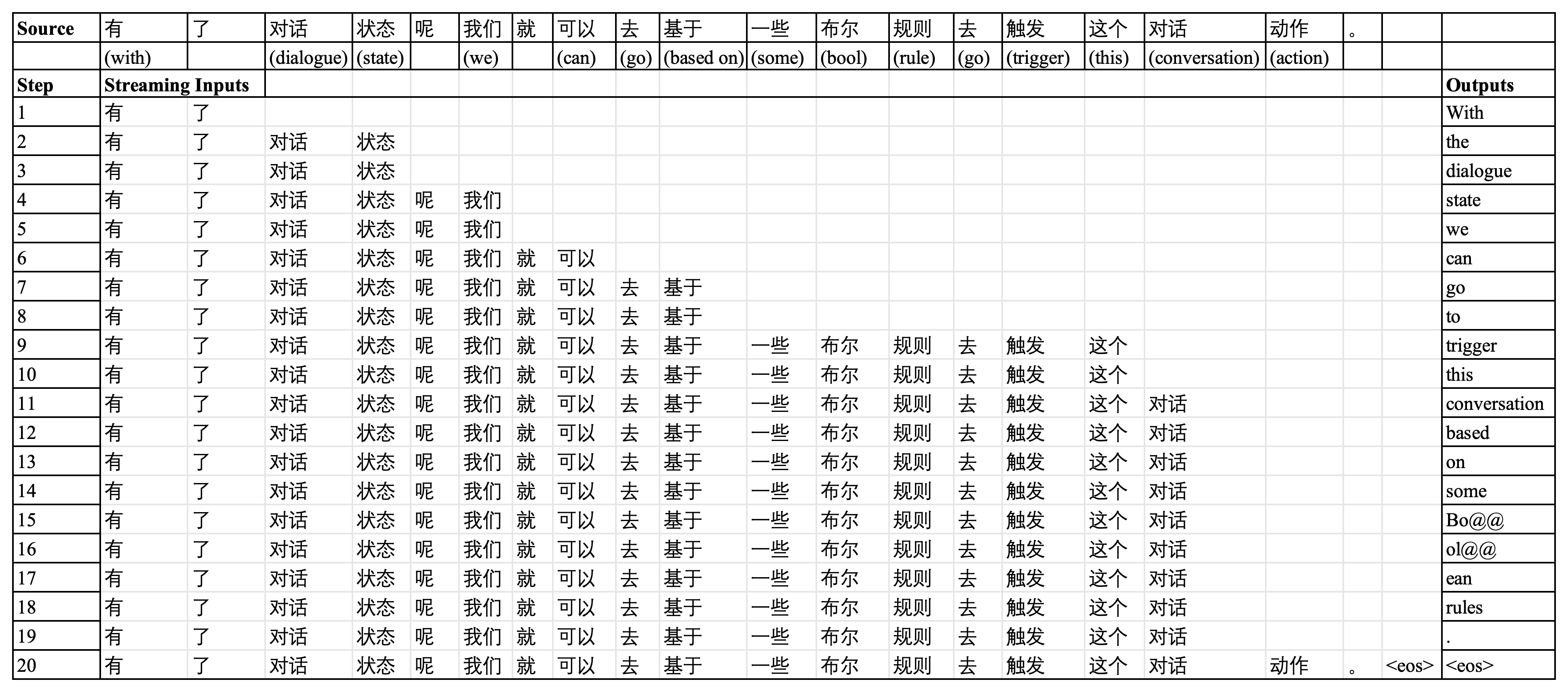}
    \caption{Case No.85 in BSTC Zh$\to$En test set, evaluated with $\lambda$ = 0.08. While there is a significant difference in word order between source and target, the PsFuture method can still make correct decisions. Specifically, at step 9, the PsFuture Policy reads an additional six tokens in sequence to ensure the accuracy of the translation.}
    \label{fig:case_study2}
\end{figure*}

Here, we present specific cases to demonstrate the effectiveness of the proposed method, as illustrated in Figure~\ref{fig:case_study1} and Figure~\ref{fig:case_study2}.
It is evident that the PsFuture policy can effectively align the source and target tokens. Even in instances where there is a significant difference in word order between source and target, the PsFuture method can still make correct decisions, waiting for more source information to proceed with the accurate translation. 

\section{Discussion on The Extra Cost Caused by Bi-directional Encoders}
During the decoding process, the use of a unidirectional encoder allows for incremental decoding, which reduces computational requirements. However, this is not feasible with bidirectional encoders. Compared to unidirectional encoders, predicting each target token necessitates the additional computation of $g(t) - 1$ encoder hidden states ($g(t)$ represents the current number of source tokens). While the extra computational load is affordable for shorter texts, it becomes considerably burdensome for longer texts, potentially imposing untenable cost. If users cannot accommodate the substantial computational demand, they can opt for a unidirectional encoder with the PsFuture method, akin to the PsFuture-W experiment mentioned in this paper which also demonstrates performance comparable to previous top non-zero-shot read/write policies.
\section{Algorithm}
\label{alg}
The inference process of PsFuture policy is summarized in Algorithm~\ref{alg:rw_policy}.

\section{Numerical Results}
The numerical main results are presented in Table~\ref{tab:main numerical results}.
\input{tables/numerical_main_results}

\input{algorithm/infer.tex}

\end{CJK}
\end{document}

%% file: figs_tex/main_result.tex
\definecolor{maincolor}{HTML}{A955FF}
\definecolor{maincolor2}{HTML}{FFA200}
\definecolor{maincolor3}{HTML}{EE005F}
\definecolor{maincolor4}{HTML}{8E3400}

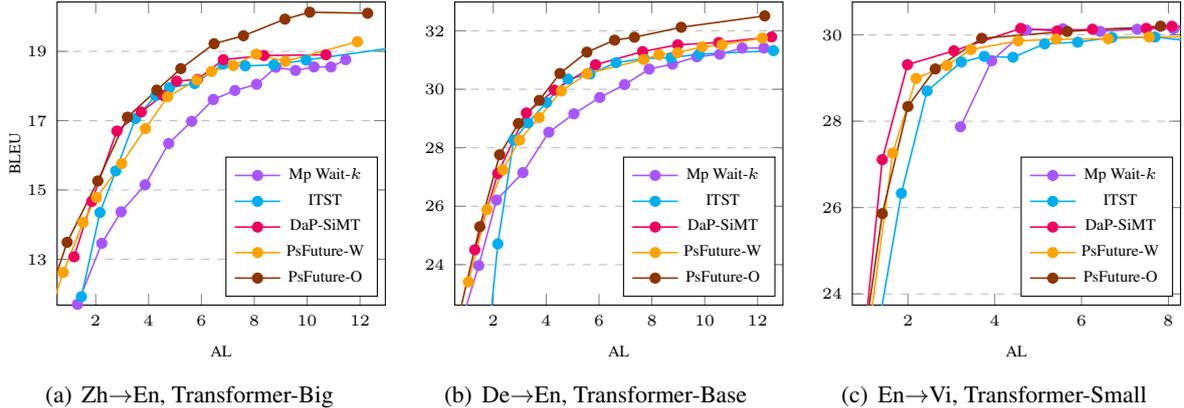
\begin{figure*}[t]
    \centering
    \pgfplotsset{width=5.9cm,height=5.6cm,
    every axis legend/.append style={at={(0.745,0.49)},anchor=north},
    every axis y label/.append style={at={(-0.08,0.5)}},
    }

    \subfigure[Zh$\to$En, Transformer-Big]{
    \begin{tikzpicture}
    \begin{axis}[
        ylabel=BLEU,
        xlabel=AL,
        enlargelimits=0.04,
        font=\tiny,
        ymajorgrids=true,
        grid style=dashed,
        legend style={font=\tiny},
        legend columns=1,
        xmin=1, xmax=12.5,
        ymin=12, ymax=20.1,
        xtick={2,4,6,8,10,12},
        ytick={13,15,17,19},
    ]
    \addplot[color=maincolor,mark=*, mark size=1.8pt,line width=0.6pt] table [y=bleu,x=al]{data/zhen/waitk.txt};
    \addplot[color=cyan,mark=*, mark size=1.8pt,line width=0.6pt] table [y=bleu,x=al]{data/zhen/itst.txt};
    \addplot[color=maincolor3,mark=*, mark size=1.8pt,line width=0.6pt] table [y=bleu,x=al]{data/zhen/dapst_max_conti1024.txt};
    \addplot[color=maincolor2,mark=*, mark size=1.8pt,line width=0.6pt] table [y=bleu,x=al]{data/zhen/psfuture_waitk_suffix4_max_conti1024.txt};
    \addplot[color=maincolor4,mark=*, mark size=1.8pt,line width=0.6pt] table [y=bleu,x=al]{data/zhen/psfuture_offline_suffix3_max_conti1024.txt};
    \legend{Mp Wait-$k$,ITST,DaP-SiMT,PsFuture-W, PsFuture-O}
    \end{axis}
    \end{tikzpicture}}
    ~
    \subfigure[De$\to$En, Transformer-Base]{
    \begin{tikzpicture}
    \begin{axis}[
        ylabel= ,
        xlabel=AL,
        enlargelimits=0.04,
        font=\tiny,
        ymajorgrids=true,
        grid style=dashed,
        legend style={font=\tiny},
        legend columns=1,
        xmin=1, xmax=12.5,
        ymin=23, ymax=32.6,
        xtick={2,4,6,8,10,12},
        ytick={24,26,28,30,32},
    ]
    \addplot[color=maincolor,mark=*, mark size=1.8pt,line width=0.6pt] table [y=bleu,x=al]{data/deen/waitk.txt};
    \addplot[color=cyan,mark=*, mark size=1.8pt,line width=0.6pt] table [y=bleu,x=al]{data/deen/itst.txt};
    \addplot[color=maincolor3,mark=*, mark size=1.8pt,line width=0.6pt] table [y=bleu,x=al]{data/deen/dapst_max_conti4.txt};
    \addplot[color=maincolor2,mark=*, mark size=1.8pt,line width=0.6pt] table [y=bleu,x=al]{data/deen/psfuture_waitk_suffix4_max_conti4.txt};
    \addplot[color=maincolor4,mark=*, mark size=1.8pt,line width=0.6pt] table [y=bleu,x=al]{data/deen/psfuture_offline_suffix4_max_conti4.txt};
    \legend{Mp Wait-$k$,ITST,DaP-SiMT,PsFuture-W,PsFuture-O}
    \end{axis}
    \end{tikzpicture}}
    ~
    \subfigure[En$\to$Vi, Transformer-Small]{
    \begin{tikzpicture}
    \begin{axis}[
        ylabel= ,
        xlabel=AL,
        enlargelimits=0.04,
        font=\tiny,
        ymajorgrids=true,
        grid style=dashed,
        legend style={font=\tiny},
        legend columns=1,
        xmin=1, xmax=8,
        ymin=24, ymax=30.5,
        xtick={2,4,6,8},
        ytick={24,26,28,30},
    ]
    \addplot[color=maincolor,mark=*, mark size=1.8pt,line width=0.6pt] table [y=bleu,x=al]{data/envi/waitk.txt};
    \addplot[color=cyan,mark=*, mark size=1.8pt,line width=0.6pt] table [y=bleu,x=al]{data/envi/itst.txt};
    \addplot[color=maincolor3,mark=*, mark size=1.8pt,line width=0.6pt] table [y=bleu,x=al]{data/envi/dapst_max_conti1024.txt};
    \addplot[color=maincolor2,mark=*, mark size=1.8pt,line width=0.6pt] table [y=bleu,x=al]{data/envi/psfuture_waitk_suffix2_max_conti1024.txt};
    \addplot[color=maincolor4,mark=*, mark size=1.8pt,line width=0.6pt] table [y=bleu,x=al]{data/envi/psfuture_offline_suffix3_max_conti1024.txt};
    \legend{Mp Wait-$k$,ITST,DaP-SiMT,PsFuture-W,PsFuture-O}
    \end{axis}
    \end{tikzpicture}}
    \captionsetup{width=16cm}
    \caption{Comparison of BLUE vs. AL curves between multi-path (abbreviated as Mp) wait-k, ITST, DaP-SiMT, and our proposed PsFuture approach on three language pairs.
    PsFuture-W and PsFuture-O denote the multi-path wait-$k$ model based PsFuture method and the offline model (P2F-enhanced) based PsFuture method, respectively.
    }   
    \label{fig:main_result}
    \end{figure*} 
    

%% file: figs_tex/ablation_study1.tex
\definecolor{maincolor}{HTML}{A955FF}
\definecolor{maincolor2}{HTML}{FFA200}
\definecolor{maincolor3}{HTML}{EE005F}
\definecolor{maincolor4}{HTML}{8E3400}

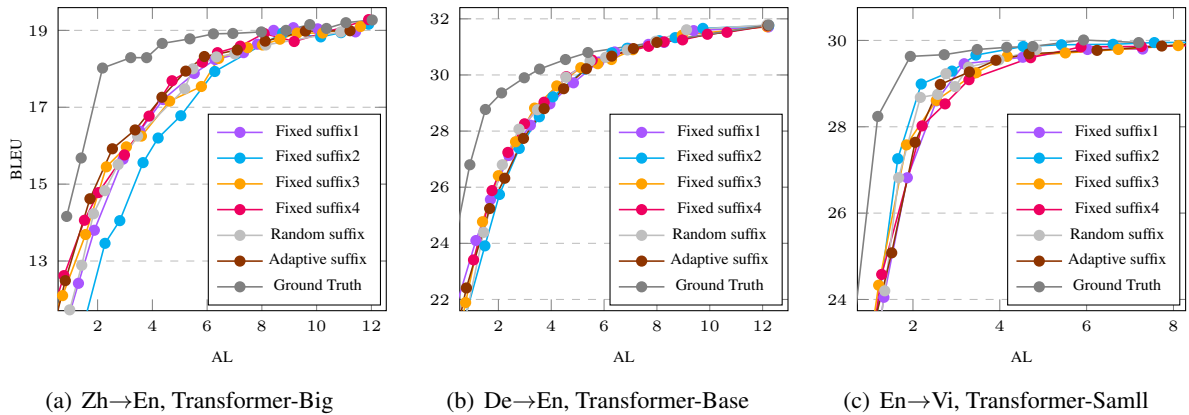
\begin{figure*}[b]
    \centering
    \pgfplotsset{width=5.9cm,height=5.6cm,
    every axis legend/.append style={at={(0.745,0.49)},anchor=north},
    every axis y label/.append style={at={(-0.08,0.5)}},
    }

\subfigure[
    Zh$\to$En, Transformer-Big
    ]{
    \begin{tikzpicture}
    \begin{axis}[
        ylabel=BLEU,
        xlabel=AL,
        enlargelimits=0.04,
        font=\tiny,
        ymajorgrids=true,
        grid style=dashed,
        legend style={font=\tiny,at={(0.72,0.65)},anchor=north},
        legend columns=1,
        xmin=1, xmax=12.1,
        ymin=12, ymax=19.3,
        xtick={2,4,6,8,10,12},
        ytick={13,15,17,19},
    ]
    \addplot[color=maincolor,mark=*, mark size=1.8pt,line width=0.6pt] table [y=bleu,x=al]{data/zhen/ablation_psfuture_waitk_suffix1_max_conti1024.txt};
    \addplot[color=cyan,mark=*, mark size=1.8pt,line width=0.6pt] table [y=bleu,x=al]{data/zhen/ablation_psfuture_waitk_suffix2_max_conti1024.txt};
    \addplot[color=maincolor2,mark=*, mark size=1.8pt,line width=0.6pt] table [y=bleu,x=al]{data/zhen/ablation_psfuture_waitk_suffix3_max_conti1024.txt};
    \addplot[color=maincolor3,mark=*, mark size=1.8pt,line width=0.6pt] table [y=bleu,x=al]{data/zhen/psfuture_waitk_suffix4_max_conti1024.txt};
    \addplot[color=lightgray,mark=*, mark size=1.8pt,line width=0.6pt] table [y=bleu,x=al]{data/zhen/ablation_psfuture_random_suffix_max_conti1024.txt};
    \addplot[color=maincolor4,mark=*, mark size=1.8pt,line width=0.6pt] table [y=bleu,x=al]{data/zhen/ablation_psfuture_waitk_adaptive_suffix_max_conti1024.txt};
    \addplot[color=gray,mark=*, mark size=1.8pt,line width=0.6pt] table [y=bleu,x=al]{data/zhen/ablation_psfuture_full_as_suffix_max_conti1024.txt};
    \legend{Fixed suffix1,Fixed suffix2,Fixed suffix3,Fixed suffix4,Random suffix, Adaptive suffix, Ground Truth}
    \end{axis}
    \end{tikzpicture}}
    ~
\subfigure[
    De$\to$En, Transformer-Base
    ]{
    \begin{tikzpicture}
    \begin{axis}[
        ylabel= ,
        xlabel=AL,
        enlargelimits=0.04,
        font=\tiny,
        ymajorgrids=true,
        grid style=dashed,
        legend style={font=\tiny,at={(0.72,0.65)},anchor=north},
        legend columns=1,
        xmin=1, xmax=12.5,
        ymin=22, ymax=32,
        xtick={2,4,6,8,10,12},
        ytick={22,24,26,28,30,32},
    ]
    \addplot[color=maincolor,mark=*, mark size=1.8pt,line width=0.6pt] table [y=bleu,x=al]{data/deen/ablation_psfuture_waitk_suffix1_max_conti4.txt};
    \addplot[color=cyan,mark=*, mark size=1.8pt,line width=0.6pt] table [y=bleu,x=al]{data/deen/ablation_psfuture_waitk_suffix2_max_conti4.txt};
    \addplot[color=maincolor2,mark=*, mark size=1.8pt,line width=0.6pt] table [y=bleu,x=al]{data/deen/ablation_psfuture_waitk_suffix3_max_conti4.txt};
    \addplot[color=maincolor3,mark=*, mark size=1.8pt,line width=0.6pt] table [y=bleu,x=al]{data/deen/psfuture_waitk_suffix4_max_conti4.txt};
    \addplot[color=lightgray,mark=*, mark size=1.8pt,line width=0.6pt] table [y=bleu,x=al]{data/deen/ablation_psfuture_random_suffix_max_conti4.txt};
    \addplot[color=maincolor4,mark=*, mark size=1.8pt,line width=0.6pt] table [y=bleu,x=al]{data/deen/ablation_psfuture_waitk_adaptive_suffix_max_conti4.txt};
    \addplot[color=gray,mark=*, mark size=1.8pt,line width=0.6pt] table [y=bleu,x=al]{data/deen/ablation_psfuture_full_as_suffix_max_conti4.txt};
    \legend{Fixed suffix1,Fixed suffix2,Fixed suffix3,Fixed suffix4,Random suffix,Adaptive suffix,Ground Truth}
    \end{axis}
    \end{tikzpicture}}
    ~
\subfigure[
    En$\to$Vi, Transformer-Samll
    ]{
    \begin{tikzpicture}
    \begin{axis}[
        ylabel= ,
        xlabel=AL,
        enlargelimits=0.04,
        font=\tiny,
        ymajorgrids=true,
        grid style=dashed,
        legend style={font=\tiny,at={(0.72,0.65)},anchor=north},
        legend columns=1,
        xmin=1, xmax=8,
        ymin=24, ymax=30.5,
        xtick={2,4,6,8},
        ytick={24,26,28,30},
    ]
    \addplot[color=maincolor,mark=*, mark size=1.8pt,line width=0.6pt] table [y=bleu,x=al]{data/envi/ablation_psfuture_waitk_suffix1_max_conti1024.txt};
    \addplot[color=cyan,mark=*, mark size=1.8pt,line width=0.6pt] table [y=bleu,x=al]{data/envi/psfuture_waitk_suffix2_max_conti1024.txt};
    \addplot[color=maincolor2,mark=*, mark size=1.8pt,line width=0.6pt] table [y=bleu,x=al]{data/envi/ablation_psfuture_waitk_suffix3_max_conti1024.txt};
    \addplot[color=maincolor3,mark=*, mark size=1.8pt,line width=0.6pt] table [y=bleu,x=al]{data/envi/ablation_psfuture_waitk_suffix4_max_conti1024.txt};
    \addplot[color=lightgray,mark=*, mark size=1.8pt,line width=0.6pt] table [y=bleu,x=al]{data/envi/ablation_psfuture_random_suffix_max_conti1024.txt};
    \addplot[color=maincolor4,mark=*, mark size=1.8pt,line width=0.6pt] table [y=bleu,x=al]{data/envi/ablation_psfuture_waitk_adaptive_suffix_max_conti4.txt};    \addplot[color=gray,mark=*, mark size=1.8pt,line width=0.6pt] table [y=bleu,x=al]{data/envi/ablation_psfuture_full_as_suffix_max_conti1024.txt};
    \legend{Fixed suffix1,Fixed suffix2,Fixed suffix3,Fixed suffix4,Random suffix,Adaptive suffix, Ground Truth}
    \end{axis}
    \end{tikzpicture}}
    \captionsetup{width=16cm}
    \caption{Effect of the pseudo-future suffix
    }   
    \label{fig:effect of the pseudo future suffix}
    \end{figure*} 
    

%% file: figs_tex/effect_of_p2f.tex
\definecolor{maincolor}{HTML}{A955FF}
\definecolor{maincolor2}{HTML}{FFA200}
\definecolor{maincolor3}{HTML}{EE005F}
\definecolor{maincolor4}{HTML}{8E3400}

\begin{figure}[h]
    \centering
    \pgfplotsset{width=0.4\textwidth,height=0.35\textwidth,
    every axis legend/.append style={at={(0.76,0.98)},anchor=north},
    every axis y label/.append style={at={(-0.08,0.5)}},
    every axis title/.append style={at={(0.13,0.92)},anchor=north,draw=black,fill=white},
    }

    \begin{tikzpicture}
    \begin{axis}[
        ylabel=BLEU,
        xlabel=AL,
        title=Zh$\to$En,
        enlargelimits=0.04,
        font=\tiny,
        grid style=dashed,
        legend style={font=\tiny,at={(0.765,0.560)},anchor=north},
        legend columns=1,
        xmin=1, xmax=12,
        ymin=13, ymax=20.3,
        xtick={2,4,6,8,10},
        ytick={13,15,17,19},
    ]
    \addplot[color=black,dashed,line width=1pt] table [y=bleu,x=al]{data/zhen/offline_mt.txt};
    \addplot[color=maincolor,mark=*, mark size=1.8pt,line width=0.6pt] table [y=bleu,x=al]{data/zhen/p2f0p0_psfuture_offline_suffix3_max_conti1024.txt};
    \addplot[color=cyan,mark=*, mark size=1.8pt,line width=0.6pt] table [y=bleu,x=al]{data/zhen/p2f0p2_psfuture_offline_suffix3_max_conti1024.txt};
    \addplot[color=maincolor2,mark=*, mark size=1.8pt,line width=0.6pt] table [y=bleu,x=al]{data/zhen/psfuture_offline_suffix3_max_conti1024.txt};
    \addplot[color=maincolor3,mark=*, mark size=1.8pt,line width=0.6pt] table [y=bleu,x=al]{data/zhen/p2f0p8_psfuture_offline_suffix3_max_conti1024.txt};
    \addplot[color=maincolor4,mark=*, mark size=1.8pt,line width=0.6pt] table [y=bleu,x=al]{data/zhen/p2f1p1_psfuture_offline_suffix3_max_conti1024.txt};
    \legend{Offline BLEU,P2F Ratio 0,P2F Ratio 0.2,P2F Ratio 0.5,P2F Ratio 0.8,P2F Ratio 1.0}
    \end{axis}
    \end{tikzpicture}

    \caption{BLEU vs. AL curves comparing among PsFuture-O experiments with varying P2F ratios. 
    }
    \label{fig:p2f}
    \end{figure}
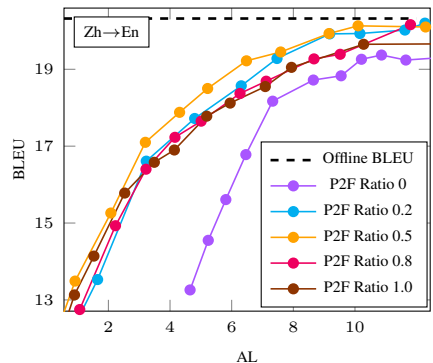

%% file: figs_tex/hallucilation_rate.tex
\definecolor{maincolor}{HTML}{A955FF}
\definecolor{maincolor2}{HTML}{FFA200}
\definecolor{maincolor3}{HTML}{EE005F}
\definecolor{maincolor4}{HTML}{8E3400}

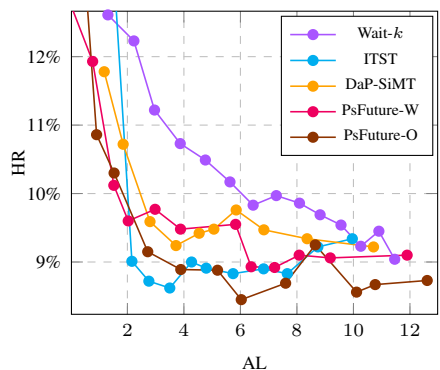
\begin{figure}[h]
    \centering
    \pgfplotsset{width=0.4\textwidth,height=0.35\textwidth,
    every axis legend/.append style={at={(0.78,0.985)},anchor=north},
    every axis y label/.append style={at={(-0.1,0.5)}},
    }
    \begin{tikzpicture}
    \begin{axis}[
        ylabel=HR,
        xlabel=AL,
        enlargelimits=0.04,
        font=\scriptsize,
        ymajorgrids=true,
        xmajorgrids=true,
        grid style=dashed,
        legend style={font=\tiny},
        legend columns=1,
        xmin=0.5, xmax=12.5,
        ymin=0.084, ymax=0.125,
        xtick={2,4,6,8,10,12},
        ytick={0.09,0.10,0.11,0.12},
        yticklabels={9\%, 10\%, 11\%, 12\%}
    ]
      
    \addplot[color=maincolor,mark=*, mark size=1.8pt,line width=0.6pt] table [y=hr,x=al]{data/zhen/hallucination_rate_waitk.txt};
    \addplot[color=cyan,mark=*, mark size=1.8pt,line width=0.6pt] table [y=hr,x=al]{data/zhen/hallucination_rate_itst.txt};
    \addplot[color=maincolor2,mark=*, mark size=1.8pt,line width=0.6pt] table [y=hr,x=al]{data/zhen/hallucination_rate_dapst.txt};
    \addplot[color=maincolor3,mark=*, mark size=1.8pt,line width=0.6pt] table [y=hr,x=al]{data/zhen/hallucination_rate_psfuture_w.txt};
    \addplot[color=maincolor4,mark=*, mark size=1.8pt,line width=0.6pt] table [y=hr,x=al]{data/zhen/hallucination_rate_psfuture_o.txt};
    \legend{Wait-$k$,ITST,DaP-SiMT,PsFuture-W,PsFuture-O}
    \end{axis}
    \end{tikzpicture}

    \caption{Hallucination Rate (HR) vs. Average Lagging (AL) curves comparing PsFuture-O with other methods.}
    \label{fig:hr}
    \end{figure}

%% file: figs_tex/ablation_study2.tex
\definecolor{maincolor}{HTML}{A955FF}
\definecolor{maincolor2}{HTML}{FFA200}
\definecolor{maincolor3}{HTML}{EE005F}
\definecolor{maincolor4}{HTML}{8E3400}

\begin{figure*}[h]
    \centering
    \pgfplotsset{width=5.9cm,height=5.6cm,
    every axis y label/.append style={at={(-0.08,0.5)}},
    every axis title/.append style={at={(0.13,0.92)},anchor=north,draw=black,fill=white},
    }
    \subfigure[
    Zh$\to$En, Transformer-Big
    ]{
    \begin{tikzpicture}
    \begin{axis}[
        ylabel=BLEU,
        xlabel=AL,
        enlargelimits=0.04,
        font=\tiny,
        ymajorgrids=true,
        grid style=dashed,
        legend style={font=\tiny,at={(0.61,0.40)},anchor=north},
        legend columns=1,
        xmin=1, xmax=10,
        ymin=12, ymax=19.3,
        xtick={2,4,6,8,10},
        ytick={13,15,17,19},
    ]
    \addplot[color=maincolor,mark=*, mark size=1.8pt,line width=0.6pt] table [y=bleu,x=al]{data/zhen/ablation_psfuture_waitk_suffix4_max_conti4.txt};
    \addplot[color=cyan,mark=*, mark size=1.8pt,line width=0.6pt] table [y=bleu,x=al]{data/zhen/ablation_psfuture_waitk_suffix4_max_conti6.txt};
    \addplot[color=maincolor2,mark=*, mark size=1.8pt,line width=0.6pt] table [y=bleu,x=al]{data/zhen/ablation_psfuture_waitk_suffix4_max_conti8.txt};
    \addplot[color=maincolor3,mark=*, mark size=1.8pt,line width=0.6pt] table [y=bleu,x=al]{data/zhen/psfuture_waitk_suffix4_max_conti1024.txt};
    \legend{Max Continuous READ 4,Max Continuous READ 6,Max Continuous READ 8,No Constraint}
    \end{axis}
    \end{tikzpicture}}
    ~
    \subfigure[
    De$\to$En, Transformer-Base
    ]{
    \begin{tikzpicture}
    \begin{axis}[
        ylabel= ,
        xlabel=AL,
        enlargelimits=0.04,
        font=\tiny,
        ymajorgrids=true,
        grid style=dashed,
        legend style={font=\tiny,at={(0.61,0.40)},anchor=north},
        legend columns=1,
        xmin=1, xmax=12.5,
        ymin=22, ymax=32,
        xtick={2,4,6,8,10,12},
        ytick={22,24,26,28,30,32},
    ]
    \addplot[color=maincolor,mark=*, mark size=1.8pt,line width=0.6pt] table [y=bleu,x=al]{data/deen/psfuture_waitk_suffix4_max_conti4.txt};
    \addplot[color=cyan,mark=*, mark size=1.8pt,line width=0.6pt] table [y=bleu,x=al]{data/deen/ablation_psfuture_waitk_suffix4_max_conti6.txt};
    \addplot[color=maincolor2,mark=*, mark size=1.8pt,line width=0.6pt] table [y=bleu,x=al]{data/deen/ablation_psfuture_waitk_suffix4_max_conti8.txt};
    \addplot[color=maincolor3,mark=*, mark size=1.8pt,line width=0.6pt] table [y=bleu,x=al]{data/deen/ablation_psfuture_waitk_suffix4_max_conti1024.txt};
    \legend{Max Continuous READ 4,Max Continuous READ 6,Max Continuous READ 8,No Constraint}
    \end{axis}
    \end{tikzpicture}}
    ~
    \subfigure[
    En$\to$Vi, Transformer-Small
    ]{
    \begin{tikzpicture}
    \begin{axis}[
        ylabel= ,
        xlabel=AL,
        enlargelimits=0.04,
        font=\tiny,
        ymajorgrids=true,
        grid style=dashed,
        legend style={font=\tiny,at={(0.61,0.40)},anchor=north},
        legend columns=1,
        xmin=1, xmax=8,
        ymin=24, ymax=30.5,
        xtick={2,4,6,8},
        ytick={24,26,28,30},
    ]
    \addplot[color=maincolor,mark=*, mark size=1.8pt,line width=0.6pt] table [y=bleu,x=al]{data/envi/ablation_psfuture_waitk_suffix2_max_conti4.txt};
    \addplot[color=cyan,mark=*, mark size=1.8pt,line width=0.6pt] table [y=bleu,x=al]{data/envi/ablation_psfuture_waitk_suffix2_max_conti6.txt};
    \addplot[color=maincolor2,mark=*, mark size=1.8pt,line width=0.6pt] table [y=bleu,x=al]{data/envi/ablation_psfuture_waitk_suffix2_max_conti8.txt};
    \addplot[color=maincolor3,mark=*, mark size=1.8pt,line width=0.6pt] table [y=bleu,x=al]{data/envi/psfuture_waitk_suffix2_max_conti1024.txt};
    \legend{Max Continuous READ 4,Max Continuous READ 6,Max Continuous READ 8,No Constraint}
    \end{axis}
    \end{tikzpicture}}
    \captionsetup{width=16cm}
    \caption{Effect of the constraint on the maximum number of continuous read}   
    \label{fig:max conti read}
    \end{figure*} 
    

%% file: tables/numerical_main_results.tex
\begin{table*}[h]
\centering
\setlength{\tabcolsep}{4pt}
\small
\begin{tabular}{c|cc|cc|cc|cc|cc}
\toprule
& \multicolumn{10}{c}{\textit{\textbf{Main Results (Figure~\ref{fig:main_result})}}}  \\
& \multicolumn{2}{c}{Mp Wait-$k$} & \multicolumn{2}{c}{ITST} & \multicolumn{2}{c}{DaP-SiMT}  & \multicolumn{2}{c}{PsFuture-W} & \multicolumn{2}{c}{PsFuture-O}  \\
\multirow{8}{*}{Zh$\to$En}  & AL  & BLEU      & AL  & BLEU         & AL  & BLEU   & AL  & BLEU  & AL  & BLEU \\
 & 1.31 & 11.7 & 0.7 & 8.91 & 1.18 & 13.07 & 0.06 & 10.99 & 0.31 & 12.12  \\
 & 2.23 & 13.46 & 1.46 & 11.92 & 1.85 & 14.67 & 0.77 & 12.62 & 0.92 & 13.49  \\
 & 2.96 & 14.37 & 2.16 & 14.35 & 2.8 & 16.7 & 1.52 & 14.06 & 2.08 & 15.26  \\
 & 3.87 & 15.15 & 2.76 & 15.55 & 3.72 & 17.25 & 2.03 & 14.78 & 3.2 & 17.1  \\
 & 4.76 & 16.34 & 3.5 & 17.06 & 4.54 & 17.73 & 2.98 & 15.76 & 4.31 & 17.88  \\
 & 5.63 & 16.98 & 4.27 & 17.72 & 5.06 & 18.14 & 3.88 & 16.77 & 5.22 & 18.5  \\
 & 6.45 & 17.61 & 4.79 & 17.95 & 5.85 & 18.19 & 4.72 & 17.69 & 6.48 & 19.22  \\
 & 7.27 & 17.87 & 5.74 & 18.07 & 6.83 & 18.76 & 5.83 & 18.17 & 7.59 & 19.45  \\
 & 8.09 & 18.05 & 6.82 & 18.63 & 8.36 & 18.88 & 6.38 & 18.42 & 9.16 & 19.93  \\
 & 8.82 & 18.54 & 7.66 & 18.58 & 10.71 & 18.9 & 7.21 & 18.59 & 10.1 & 20.13  \\
 & 9.56 & 18.45 & 8.74 & 18.61 &  &  & 8.08 & 18.92 & 12.29 & 20.1  \\
 & 10.26 & 18.55 & 9.96 & 18.75 &  &  & 9.18 & 18.71 &  &   \\
 & 10.9 & 18.55 & 13.68 & 19.15 &  &  & 11.9 & 19.28 &  &   \\
 & 11.46 & 18.76 &  &  &  &  &  &  &  &   \\
\midrule
\multirow{8}{*}{De$\to$En}  & AL  & BLEU     & AL  & BLEU         & AL  & BLEU   & AL  & BLEU  & AL  & BLEU \\
 & 0.47 & 21.08 & 1.57 & 19.2 & 0.49 & 21.65 & 1.06 & 23.41 & 1.49 & 25.3  \\
 & 1.45 & 23.97 & 2.17 & 24.71 & 1.3 & 24.51 & 1.76 & 25.88 & 2.24 & 27.76  \\
 & 2.12 & 26.21 & 2.77 & 28.26 & 2.17 & 27.12 & 2.36 & 27.24 & 2.95 & 28.83  \\
 & 3.12 & 27.15 & 3.31 & 28.85 & 3.25 & 29.19 & 2.99 & 28.26 & 3.74 & 29.62  \\
 & 4.1 & 28.53 & 4.01 & 29.55 & 4.31 & 29.97 & 3.73 & 29.03 & 4.52 & 30.54  \\
 & 5.05 & 29.16 & 4.82 & 30.35 & 5.87 & 30.84 & 4.57 & 29.94 & 5.54 & 31.27  \\
 & 6.03 & 29.72 & 5.66 & 30.52 & 7.65 & 31.29 & 5.55 & 30.53 & 6.59 & 31.68  \\
 & 6.97 & 30.16 & 6.65 & 30.91 & 8.98 & 31.52 & 7.69 & 31.02 & 7.34 & 31.78  \\
 & 7.9 & 30.69 & 7.7 & 31.05 & 10.53 & 31.6 & 8.27 & 31.18 & 9.11 & 32.12  \\
 & 8.78 & 30.86 & 8.73 & 31.08 & 12.53 & 31.79 & 8.97 & 31.25 & 12.27 & 32.51  \\
 & 9.7 & 31.11 & 9.79 & 31.2 &  &  & 9.91 & 31.45 &  &   \\
 & 10.57 & 31.2 & 12.6 & 31.32 &  &  & 10.65 & 31.52 &  &   \\
 & 11.42 & 31.41 &  &  &  &  & 12.18 & 31.75 &  &   \\
 & 12.24 & 31.41 &  &  &  &  &  &  &  &   \\
\midrule
\multirow{8}{*}{En$\to$Vi}  & AL  & BLEU          & AL  & BLEU         & AL  & BLEU   & AL  & BLEU & AL  & BLEU \\
 & 3.21 & 27.87 & 1.29 & 23.06 & 0.89 & 21.89 & 0.84 & 21.16 & 0.21 & 18.08  \\
 & 3.93 & 29.4 & 1.85 & 26.33 & 1.41 & 27.11 & 1.65 & 27.26 & 0.86 & 21.96  \\
 & 4.73 & 30.11 & 2.44 & 28.7 & 1.99 & 29.31 & 2.19 & 28.99 & 1.41 & 25.86 \\
 & 5.57 & 30.14 & 3.23 & 29.37 & 3.06 & 29.63 & 2.9 & 29.29 & 2 & 28.34  \\
 & 6.43 & 30.08 & 3.76 & 29.5 & 4.6 & 30.15 & 3.45 & 29.66 & 2.63 & 29.21  \\
 & 7.28 & 30.13 & 4.42 & 29.48 & 5.44 & 30.09 & 4.54 & 29.86 & 3.7 & 29.92  \\
 & 8.12 & 30.14 & 5.15 & 29.79 & 6.25 & 30.13 & 5.42 & 29.9 & 5.67 & 30.08  \\
 & 8.93 & 30.11 & 5.91 & 29.83 & 7.49 & 30.15 & 6.61 & 29.91 & 7.82 & 30.2  \\
 & 9.7 & 30.1 & 6.7 & 29.94 & 8.08 & 30.2 & 7.56 & 29.95 & 9.91 & 30.14  \\
 & 10.43 & 30.2 & 7.69 & 29.95 & 8.74 & 30.17 & 9.1 & 29.96 &  &   \\
 & 11.13 & 30.16 & 8.67 & 29.84 & 9.61 & 30.01 &  &  &  &   \\
 & 11.79 & 30.13 & 9.93 & 29.95 & 10.67 & 30.11 &  &  &  &   \\
 & 12.41 & 30.16 & 12.58 & 30.01 & 11.69 & 30.1 &  &  &  &   \\
 & 13.01 & 30.18 &  &  &  &  &  &  &  &   \\

\bottomrule

\end{tabular}
\caption{Numerical results in Figure~\ref{fig:main_result}. }
\label{tab:main numerical results}
\end{table*}

%% file: algorithm/infer.tex
\newcommand\mycommfont[1]{\small\ttfamily\textcolor{blue}{#1}}
\SetCommentSty{mycommfont}

\SetKwInput{KwInput}{Input}                
\SetKwInput{KwOutput}{Output}              
\SetKwInput{Init}{Init}
\SetKw{Continue}{continue}
\SetKw{or}{or}
\SetKw{and}{and}
\SetKwFunction{Score}{score}
\SetKwFunction{Last}{last}
\SetKwFunction{Add}{add}
\SetKwFunction{append}{Append}
\SetKwFunction{AddAll}{add\_all}
\SetKwFunction{Top}{top}
\SetKwFunction{Max}{max}
\SetKwFunction{Len}{len}
\DontPrintSemicolon
\begin{algorithm}[h]
  \caption{SiMT inference with PsFuture policy}
  \label{alg:rw_policy}
  \KwInput{streaming source tokens: $\mathbf{X}_{\leq j}$,\newline
    threshold: $\delta$,\newline
    target idx: $i \leftarrow 1$,\newline 
    source idx: $j \leftarrow 2$,\newline 
    max continuous READ constraint: $r_{max}$,\newline
    current number of continuous READ: $r_{c} \leftarrow 1$
  }
  \KwOutput{target tokens: $\mathbf{Y} \leftarrow$ \{<$\mathtt{BOS}$>\}}
  \While{$\mathbf{Y}_{i-1} \neq$ <$\mathtt{EOS}$>}{
  calculate R/W confidence (cosine distance) $c$ with $\mathbf{Y}_{i-1}$ using the PsFuture policy mentioned in \ref{sec:psfuture method};
  
  \If{$c \leq \delta$ \or $r_{c} \geq r_{max}$} {
  translate $y_i$ with $\mathbf{X}_{\leq j}$,$\mathbf{Y}_{\leq i-1}$;

  {\If{$y_{i} \neq$ <$\mathtt{EOS}$> \or $j \geq |\mathbf{X}|$}{
  \tcp{execute WRITE action}

  $\mathbf{Y}.\append(y_i)$;
  
  $r_{c} \leftarrow 0$;
  
  $i \leftarrow i+1$;}
  }
  \Else{\tcp{execute READ action}
  
  $j \leftarrow j+1$;
  
  $r_{c} \leftarrow r_{c}+1$;
  }}
  \Else{\tcp{execute READ action}
  
  $j \leftarrow j+1$;
  
  $r_{c} \leftarrow r_{c}+1$;
  }
  }
  \Return $\mathbf{Y}$
\end{algorithm}